\documentclass[conference]{IEEEtran}
\IEEEoverridecommandlockouts
\usepackage{cite}
\usepackage{amsmath,amssymb,amsfonts}
\usepackage{algorithmic}
\usepackage{graphicx}
\usepackage{pgfplots}
\usepackage{stfloats}
\usepackage{subfig}
\usepackage{textcomp}
\usepackage{xcolor}
\def\BibTeX{{\rm B\kern-.05em{\sc i\kern-.025em b}\kern-.08em
    T\kern-.1667em\lower.7ex\hbox{E}\kern-.125emX}}

\usepackage{hyperref}
\usepackage[T1]{fontenc} 
\usepackage[utf8]{inputenc}
\usepackage{verbatim}
\usepackage{float}
\usepackage{multirow}
\usepackage{booktabs}
\usepackage{tikz}

\usepgfplotslibrary{groupplots}
\usetikzlibrary{matrix,positioning}
\pgfplotsset{compat=newest}

\begin{document}

\title{Towards Explaining Monte-Carlo Tree Search \\by Using Its Enhancements
\thanks{This article is based on the work of COST Action CA22145 -- GameTable, supported by COST (European Cooperation in Science and Technology).
\\
\indent This research was supported in part by the National Science Centre, Poland, under project number 2021/41/B/ST6/03691 (Jakub Kowalski). 
}
}

\author{\IEEEauthorblockN{Jakub Kowalski\IEEEauthorrefmark{1}, Mark H. M. Winands\IEEEauthorrefmark{2}, Maksymilian Wiśniewski\IEEEauthorrefmark{1}, 
Stanis{\l}aw Reda\IEEEauthorrefmark{1},
Anna Wilbik\IEEEauthorrefmark{2}}

\IEEEauthorblockA{\IEEEauthorrefmark{1}\textit{Faculty of Mathematics and Computer Science, University of Wroc{\l}aw}\\jakub.kowalski@cs.uni.wroc.pl, \{330939, 330007\}@uwr.edu.pl}
\IEEEauthorrefmark{2}\textit{Department of Advanced Computing Sciences, Maastricht University}\\
\{m.winands, a.wilbik\}@maastrichtuniversity.nl
}

\maketitle

\begin{abstract}
Typically, research on Explainable Artificial Intelligence (XAI) focuses on black-box models within the context of a general policy in a known, specific domain. This paper advocates for the need for knowledge-agnostic explainability applied to the subfield of XAI called Explainable Search, which focuses on explaining the choices made by intelligent search techniques. It proposes Monte-Carlo Tree Search (MCTS) enhancements as a solution to obtaining additional data and providing higher-quality explanations while remaining knowledge-free, and analyzes the most popular enhancements in terms of the specific types of explainability they introduce. So far, no other research has considered the explainability of MCTS enhancements. We present a proof-of-concept that demonstrates the advantages of utilizing enhancements.

\end{abstract}



\section{Introduction}

With the increase in the amount and variety of AI systems surrounding us and substantially affecting our lives, the need 
to understand the reasons behind their decisions has also naturally increased.
Thus, the growth of Explainable Artificial Intelligence (XAI), as a field of AI research responsible for enhancing intelligent agents with the ability to describe their decision process in a human-understandable way, hoping it would increase the trust of society \cite{BARREDOARRIETA202082}.

However, XAI research is focused mainly on explaining black-box (usually machine-learning) models, in the context of a general policy, within a known, specific domain \cite{guidotti2018survey,wrede2022linguistic}. 
In this paper, we would like to introduce and justify the entirely opposite concept. We provide a breakdown of the proposed methodology and preliminary results towards its realization.

Explainable Search (XS) is a subfield of XAI that concentrates on explaining choices made by intelligent search techniques, which usually construct an entire plan of actions, often within domains requiring sequential decision making \cite{baier2020explainable}.
Still, even when applied to knowledge-agnostic algorithms like Monte-Carlo Tree Search (MCTS)  \cite{mcts2}, XS tends to heavily leverage the available domain knowledge.

This is a natural tendency, given how a relatively low amount of information, and mostly hard to turn into a human-readable form, can be decoded from the search tree itself.
Our proposition to solve this problem, never stated before, is to take advantage of the numerous MCTS enhancements.
Techniques such as, e.g., MCTS Solver \cite{winands08b}, MAST \cite{finnsson2008simulation}, GRAVE \cite{cazenave2015generalized}, NST \cite{tak2012}, and PN-MCTS \cite{Kowalski2024ProofNumber} are still domain-independent, and can be successfully applied to unknown problems, yet each of them adds a unique amount of knowledge, allowing for better explanation of both the AI decision and the problem domain itself.

In this paper, we describe how each of the enhancements can be used to improve the explainability of MCTS, and argue about the importance of a secondary goal of XAI -- domain explanation.
We believe that gathering knowledge to explain the behavior of the algorithm is also a great opportunity for improving the user's understanding of the problem domain itself.
This, in turn, may pay off, as such information can then be  used for designing better algorithms, this time taking advantage of the obtained domain knowledge.

We present the results of our initial experiments, using the described methodology to explain the decisions of the MCTS-based agent that can be applied to over 1,400 games available in the Ludii General Game Playing System \cite{Piette2020Ludii}.

\section{Related Research} 

Explainable search  \cite{baier2020explainable} and explainability of MCTS is still a relatively new topic, and only a handful of approaches have been presented in the literature. Introducing MCTS as a target for explainable search and describing elements of explainable MCTS has been proposed by \cite{baier2021towards}.
The method of explaining MCTS via answering logic-based queries regarding the state of the search tree was proposed by \cite{an2024enabling}.
An analysis of MCTS explanations types set in a two-dimensional space, containing features (statistics, context) and tree scope (flat, siblings, sequence), has been tested on SameGame by \cite{sironi2023explainable}.
An approach to explain Sudoku solutions by a custom depth-first branch-and-bound-like search with  understandability put as an element of the cost function has been presented in \cite{YngviExSudoku}. 
Another branch focuses on specialized Large Language Models (LLMs), either as a means to allow more natural queries for MCTS \cite{an2025combining}, or to use MCTS for enhancing the reasoning capabilities of LLMs \cite{gao2024interpretable}.

Most of the referenced works focus on single-player games and known domains.
As our goal is to approach explainability from the perspective of general-purpose, knowledge-agnostic usage, the best domain for such endeavors is General Game Playing (GGP).
GGP systems provide a means for AI agents to play any game from the database, given either raw encoding of its rules or a forward model, as it is a game not seen before by the agent.

\section{Knowledge-agnostic Explainable Search}

\subsection{Explainable problem domains}

The most common understanding of the XAI goal is to make human users understand the reasons justifying the decision made by an AI agent.
However, to convince the user that the decision made by the agent is rational and well-grounded, the explanations need to rely on the user's knowledge about the problem and take into account their level of expertise. 
Yet we may easily imagine an alternative scenario, where the user observes an agent play to understand the problem, rather than to understand the agent.

In such a scenario, the responses of XS should be tailored accordingly. 
More emphasis should be put on the context of the agent's decisions, and highlighting the properties of the situation that were crucial for decision making.
Understanding what properties these are, how they generalize, and what type of response is best to handle them is key to understanding the problem domain by the user.

In the game domain, the most common example is Chess learning, with various AI-based systems that analyze players' games in the context of AI-proposed moves and point out mistakes and how to avoid them. 
Although these systems are far from perfect, and their usability is limited, they are still a popular and useful tool for players of wide range of skill levels.

\subsection{Explainability of unknown games}

However, only a few board games are as thoroughly analyzed as chess. For most of the other known games, only some level of AI-based expertise is available. However, many lesser-known games exist, e.g., archeological or newly designed, where no knowledge at all is available.

In the presence of the Ludii system \cite{Piette2020Ludii}, containing a database with over 1,400 games useful to both AI and archeological research, there is a pressing need for algorithms that not only play well, but also teach humans how to play \cite{Soemers2024GameTableWorking}.
Although if given access to encoded game rules, some knowledge extraction can be done \cite{9619052}, but it is neither perfect nor possible when only a forward model is available.

Understanding an unknown problem via explainable AI requires knowledge-agnostic algorithms. 
Currently, the most popular algorithm meeting this requirement, widely applicable to games, but not only, is MCTS \cite{swiechowski2023monte}.
Another advantage of MCTS is that many of its well-established enhancements are also knowledge-free and can be advantageously used for improving the explainability of the performed search.



\newcommand{\imw}[0]{30mm}
\begin{table*}[!htb]\renewcommand{\arraystretch}{1.05}\small
\newcommand{\colt}[1]{\multicolumn{2}{c|}{#1}}
\newcommand{\coltt}[1]{\multicolumn{3}{c|}{#1}}
\newcommand{\colttI}[1]{\multicolumn{3}{c||}{#1}}
\newcommand{\coltttt}[1]{\multicolumn{5}{c|}{#1}}
\newcommand{\tb}[1]{\textbf{#1}}
\newcommand{\nb}[1]{#1}
\caption{Example game positions and moves with explanations.}\label{tab:examples}
{\footnotesize 
\begin{center}\begin{tabular}{|p{\imw}|p{75mm}|p{75mm}|}\hline
\textbf{Position} & \textbf{Data} & \textbf{Explanation} \\\hline

\raisebox{-\totalheight}{\includegraphics[width=\imw]{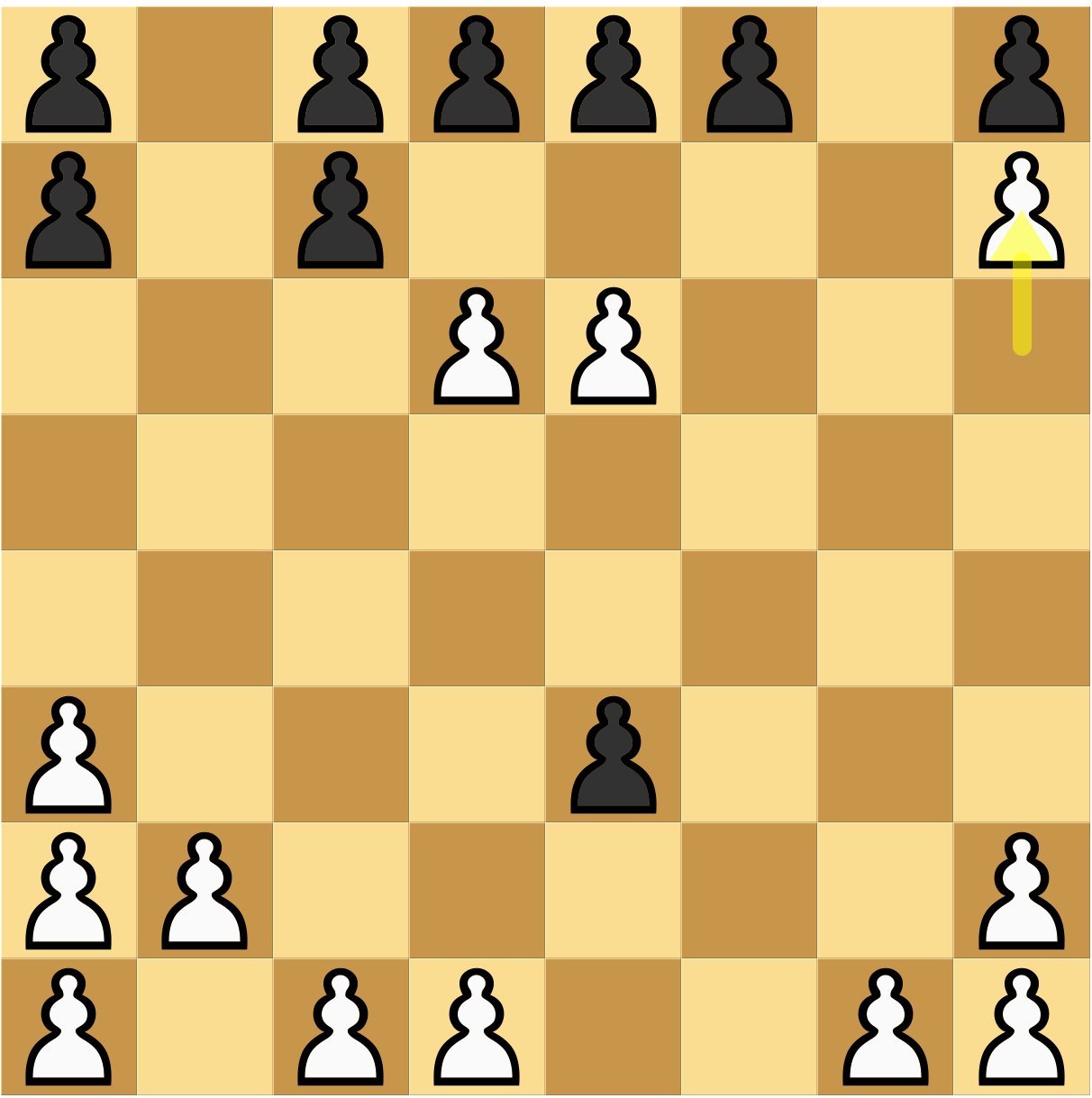}}\vspace{2px}
&
\textbf{Breakthrough (white)}: \textbf{Score bounded} \newline
Performed 1599 iterations. Previous turn score: 0.3072.\newline 
Selected node:\newline 
{move: H6-H7, visits: 204, score: 0.4608, solved node with score 1.0000 (win)}\newline 
Other nodes:\newline 
{move: B2-C3, visits: 17, score: -0.2941, pess: -1.0000, opt: 1.0000}\newline 
{move: D6-C7, visits: 155, score: 0.3935, pess: -1.0000, opt: 1.0000}... 
&
There are 23 moves available: 1 with decisive advantage (proven win), 15 with slight advantage (above 55.56\%), 5 balanced ($\sim$50\%), 2 with slight disadvantage (below 35.29\%).\newline 
Selected move, H6-H7, leads to a proven win in 2 turns. After we play this move, the most probable sequence of following moves will be: E3-E2, H7-G8.\newline 
\\\hline

\raisebox{-\totalheight}{\includegraphics[width=\imw]{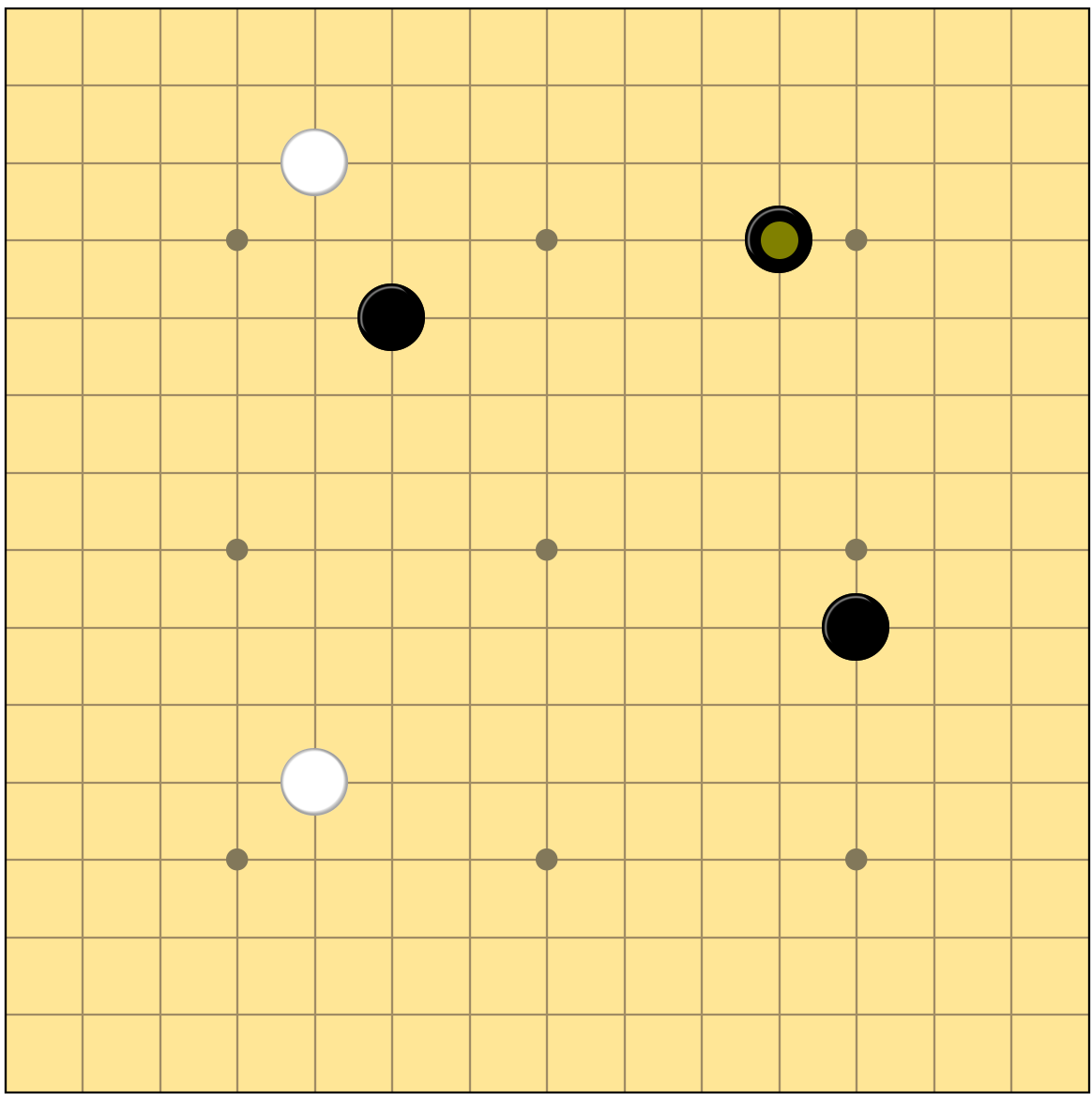}}\vspace{2px}
&
\textbf{Gomoku (black)}: \textbf{Score bounded, GRAVE} \newline
Performed 4671 iterations. Previous turn score: 0.0980.\newline 
Selected node:\newline 
{move: K12+Marker1, visits: 1111, score: 0.0981, AMAF visits: 1923, AMAF score: 0.1014, pess: -1.0000, opt: 1.0000}\newline 
Other nodes:\newline 
{move: E12+Marker1, visits: 95, score: 0.0316, AMAF visits: 1204, AMAF score: 0.0880, pess: -1.0000, opt: 1.0000}\newline 
{move: K14+Marker1, visits: 5, score: -0.6000, AMAF visits: 1090, AMAF score: 0.0165, pess: -1.0000, opt: 1.0000}...
&
There are 221 moves available: 89 with decisive advantage (highly likely win), 5 with slight advantage (above 55.03\%), 22 balanced ($\sim$50\%), 21 with slight disadvantage (below 43.75\%), 84 with decisive disadvantage (highly likely loss).
Selected move: K12+Marker1.
Our position is balanced (estimated win probability: 54.91\%).
111 of alternative moves are significantly worse. 84 of them are highly likely a defeat.
The selected best move, K12+Marker1, has estimated win probability of 54.91\%, but it was not chosen based on that metric. There are 95 moves with higher win probability (best of them, M14+Marker1, is better by 45.09\%). However, these moves have slightly worse AMAF scores (2.94\% worse for M14+Marker1), which influenced the result. 
\\\hline

\raisebox{-\totalheight}{\includegraphics[width=\imw]{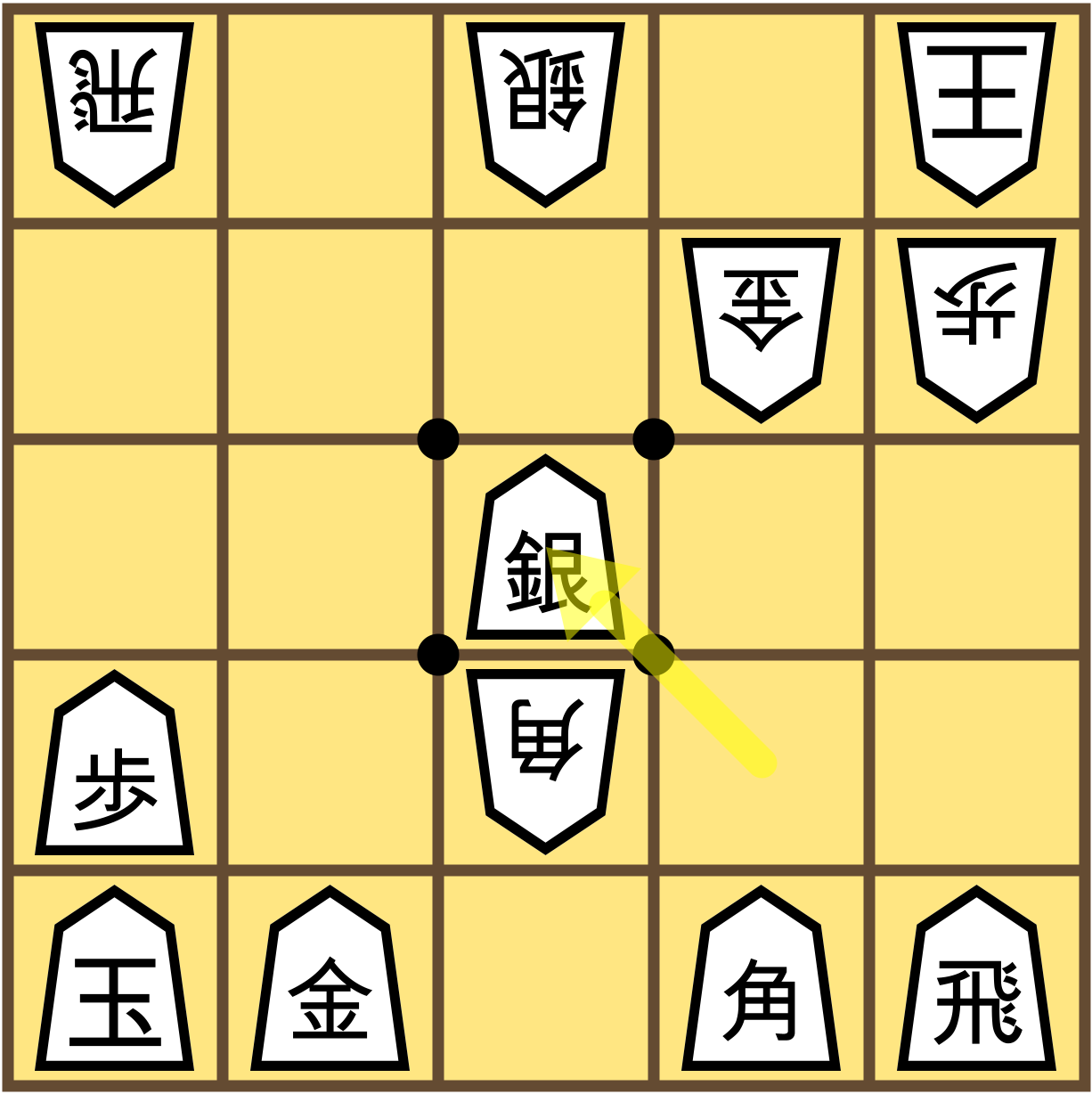}}\vspace{2px}
&
\textbf{MiniShogi (bottom)}: \textbf{Score bounded} \newline
Performed 206 iterations. Previous turn score: 0.1250. \newline 
Selected node: \newline 
{move: D2-C3, visits: 31, score: 0.3548, pess: -1.0000, opt: 1.0000} \newline 
Other nodes: \newline 
{move: D1-E2, visits: 11, score: -0.0909, pess: -1.0000, opt: 1.0000} \newline 
{move: D2-C1, visits: 14, score: 0.0000, pess: -1.0000, opt: 1.0000} \newline 
{move: B1-B2, visits: 28, score: 0.2857, pess: -1.0000, opt: 1.0000} \newline 
{move: E1-E3, visits: 7, score: -0.4286, pess: -1.0000, opt: 1.0000} \newline 
{move: B1-C2, visits: 17, score: 0.0588, pess: -1.0000, opt: 1.0000} \newline 
{move: D2-E3, visits: 8, score: -0.2500, pess: -1.0000, opt: 1.0000} \newline 
...
&
There are 14 moves available: 4 with slight advantage (above 55.56\%), 6 balanced ($\sim$50\%), 3 with slight disadvantage (below 37.50\%), 1 with decisive disadvantage (highly likely loss). \newline 
Selected move: D2-C3. \newline 
Our position is slightly advantageous (estimated win probability: 67.74\%). \newline 
The overall estimation of our position improved over the previous turn (11.49\% increased win probability). \newline 
11 of alternative moves are significantly worse. 1 of them is highly likely a defeat.
The selected move is slightly better than all other options (3.46\% increased win probability over the next best option, B1-B2).  
\\\hline

\raisebox{-\totalheight}{\includegraphics[width=\imw]{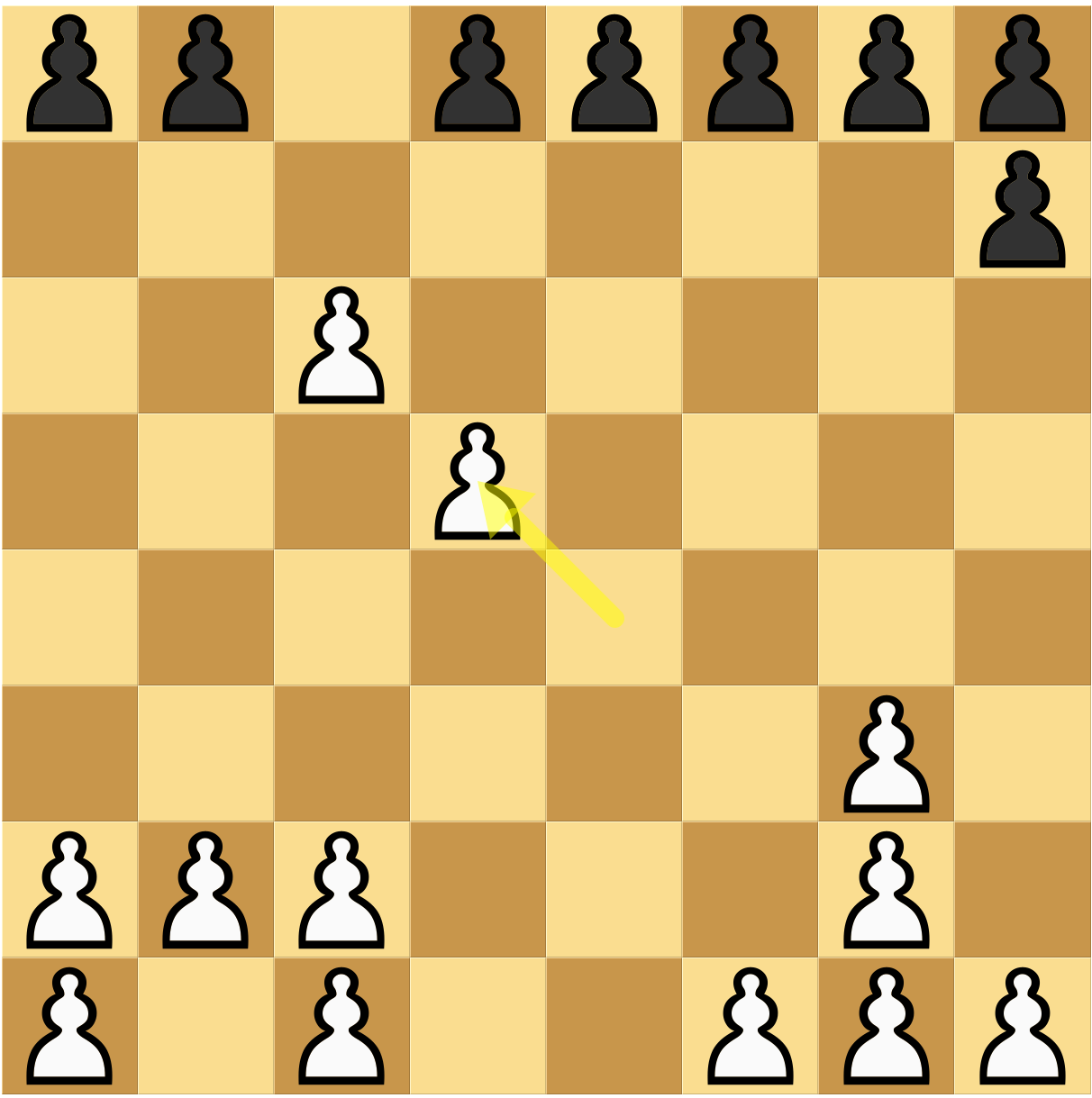}}\vspace{2px}
&
\textbf{Breakthrough (white)}: \textbf{Score bounded, MAST, NST} \newline
Performed 7351 iterations. Previous turn score: 0.5612.\newline
Selected node:\newline
{move: E4-D5, visits: 5146, score: 0.7482, 1-gram visits: 30580, 1-gram score: 0.275278, 2-gram visits: 5411, 2-gram score: 0.719830, pess: -1.0000, opt: 1.0000}\newline
Other nodes:\newline
{move: C6-B7, visits: 197, score: 0.4822, 1-gram visits: 13656, 1-gram score: 0.499414, 2-gram visits: 317, 2-gram score: 0.476341, pess: -1.0000, opt: 1.0000}...
% ...
&
There are 25 moves available: 1 with decisive advantage (above 87.41\%), 24 with slight advantage (above 56.82\%). \newline
Our position is generally advantageous (the estimated win probability for the worst of available moves is 56.82\%). \newline
The selected move, E4-D5, is significantly better than all other options (13.30\% increased win probability over the next best option, C6-B7). \newline
3 moves are significantly better (at least 11.21\% better) than the selected one according to the MAST metric. One move (the selected one) is significantly better (at least 85.99\%) than the rest according to the NST(2) metric. 
\\\hline

\raisebox{-\totalheight}{\includegraphics[width=\imw]{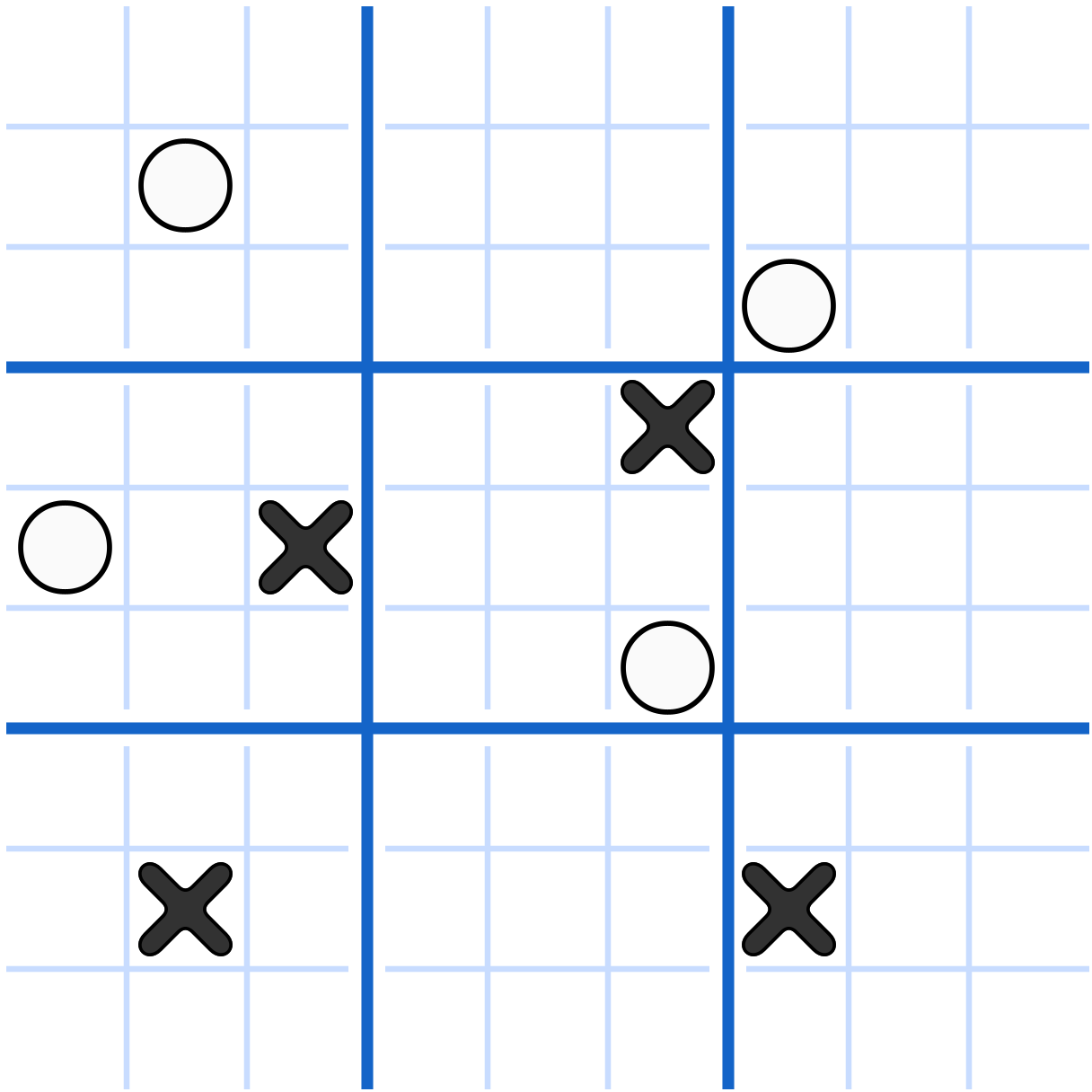}}\vspace{2px}
&
\textbf{Ultimate Tic-Tac-Toe (white): Score bounded } \newline
Performed 24465 iterations. Previous turn score: 0.0795. \newline
Selected node: \newline
{move: I5+Disc1, visits: 4922, score: 0.0809, pess: -1.0000, opt: 1.0000}\newline
Other nodes:\newline
{move: H5+Disc1, visits: 1122, score: 0.0098, pess: -1.0000, opt: 1.0000}\newline
{move: G4+Disc1, visits: 2749, score: 0.0589, pess: -1.0000, opt: 1.0000}...
&
There are 9 moves available: 9 balanced (~50\%).\newline
Selected move: I5+Disc1.\newline
Our position is balanced (estimated win probability: 54.04\%).\newline
The selected best move, I5+Disc1, has estimated win probability of 54.04\%, but it was not chosen based on that metric. There is one move (G5+Disc1) with higher win probability, which is better by 0.02\%). However, this move has worse visit count, which influenced the result. 
\\\hline

\raisebox{-\totalheight}{\includegraphics[width=\imw]{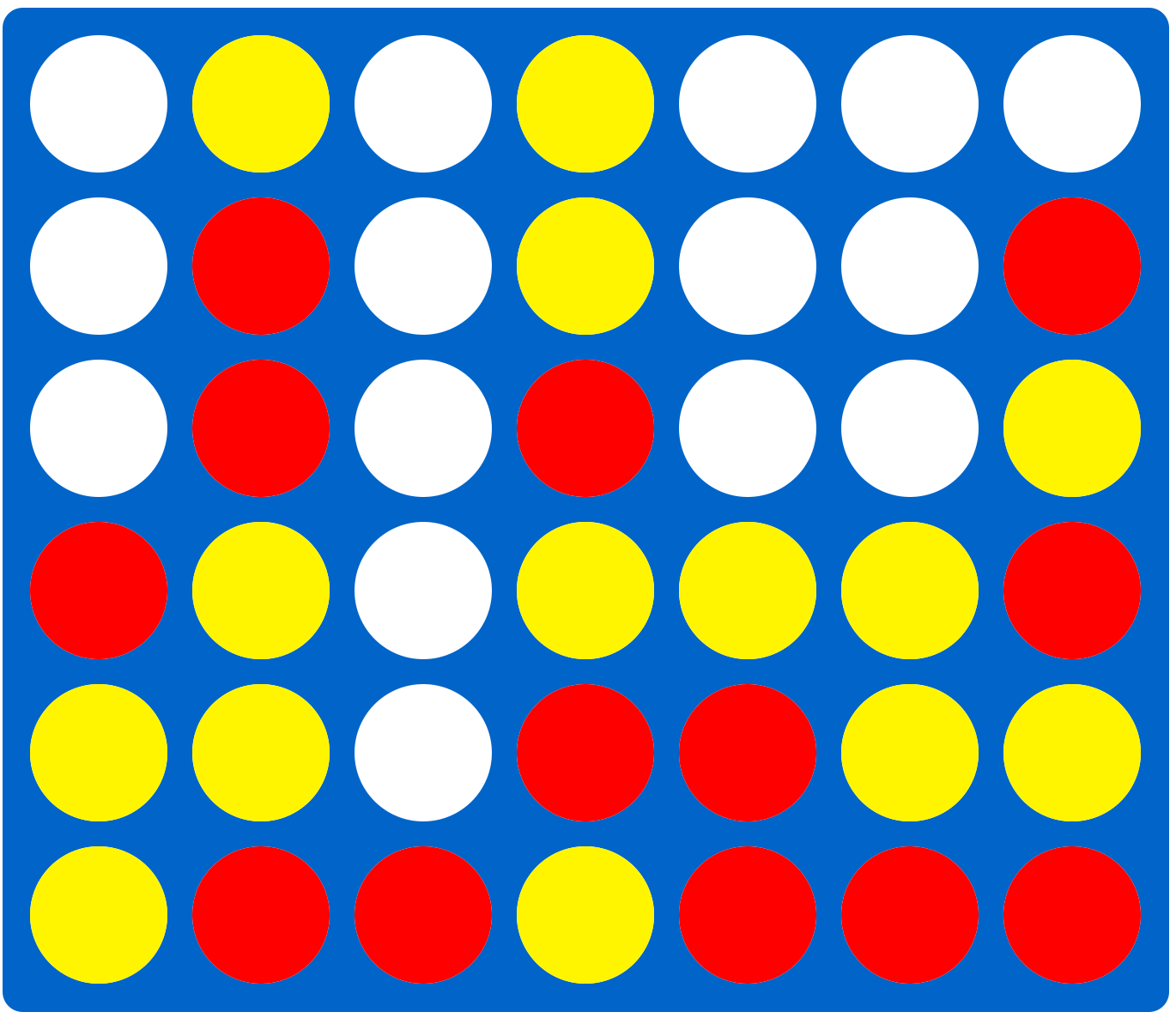}}\vspace{2px}
&
\textbf{Connect Four (yellow)}: \textbf{Score bounded} \newline
Performed 0 iterations. Previous turn score: 0.3352.\newline
Selected node:\newline
{move: E1/2+Disc1, visits: 64, score: 0.5938, solved node with score 1.0000 (win)}\newline
Other nodes:\newline
{move: A1/3+Disc1, visits: 10, score: 0.0000, pess: -1.0000, opt: 1.0000}\newline
{move: C1/1+Disc1, visits: 5, score: 0.2000, solved node with score -1.0000 (loss)}...
&
There are 5 moves available: 1 with decisive advantage (proven win), 1 with slight advantage (above 70.00\%), 1 balanced ($\sim$50\%), 1 with slight disadvantage (below 33.33\%), 1 with decisive disadvantage (proven loss).\newline
Selected move, E1/2+Disc1, leads to a proven win in 2 turns. After we play this move, the most probable sequence of following moves will be: A1/3+Disc2, E1/3+Disc1.\newline
4 of alternative moves are significantly worse. 1 of them is a proven defeat.\newline
\\\hline

\end{tabular}
\end{center}
}
\end{table*}

\subsection{Explainability of MCTS enhancements}

In this section, we would like to briefly introduce some of the notable knowledge-agnostic MCTS enhancements and provide a high-level overview of the additional explainable knowledge that is possible to obtain from them.

The Move-Average Sampling Technique (MAST) \cite{finnsson2008simulation} involves storing average rewards for all simulated actions, regardless of the game state. These statistics are used to guide the simulation policy.
This additional data potentially allows us to explain a move chosen by the algorithm in the context of its average performance. It might be interesting for the user to know if e.g., the performed move is good in general.
Or even more for the opposite, that the move was chosen despite being worse than the alternatives, if taking into account only contextless MAST data.

N-Gram Selection Technique (NST) \cite{tak2012}, is an extension of MAST that takes into account longer move sequences. In particular, NST storing sequences of length two allows taking into account the ``average good reply''.
Here, we can extend the contextual explanation of the move by describing the decision taken from the perspective of move exchange and replies, providing information on how in line or in contrast it was with NST recommendations.

Rapid Action Value Estimation (RAVE)  and Generalized RAVE (GRAVE) \cite{cazenave2015generalized} also compute the reward estimation for moves, but it is based on All-Moves-As-First (AMAF) heuristics \cite{Gelly07}, and there are several differences compared to MAST. Move scores are updated for moves played in both selection and simulation phases within the subtree of a node. These values are used to bias the selection.
Thus, if the algorithm uses RAVE for actual play, it influences the choice of the move, and provides an additional type of explanation.
In particular, we may deduce that, e.g., the played move was chosen because of RAVE value, while otherwise some other move would be used, and this one was $k$-th in order.

Score bounded MCTS \cite{cazenave2011ScoreBoundedMCTS} backpropagates information from leaf nodes that are terminal states, and for each subtree stores pessimistic and optimistic achievable rewards.
This allows us to explain trees and moves in terms of losing and winning positions. In particular, we can reliably mark a position with forced moves when it clearly emerges with alternative moves leading to inevitable doom.
Also, the amount of solved subtrees within our search tree influences the uncertainty of our search result, and simultaneously, the uncertainty of our explanation.

Proof-Number Search (PNS) \cite{pns2} is a tree-solving technique that can play a similar role to the Score bounded extension \cite{Kowalski2024ProofNumber}, but instead of bounds, it collects information about estimated workload to prove or disprove each subtree.
For the sake of explainability, we may, e.g., use the balance between a potential work to solve the tree in favor of the player and that of the opponent. 
If a strong imbalance occurs, this provides an additional insight about the game situation, not simply derived from the search node values.

Although more MCTS enhancements can be potentially used
(e.g., Last Good Reply \cite{baier2010}, Transposition Tables \cite{childs2008}), the above list covers the most popular ones, and the ones with a clear potential to improve explainability.

\section{Experiments}\label{sec:agents}


As for the proof of concept, we have implemented explainable MCTS with enhancements for the Ludii system \cite{Piette2020Ludii}. We used Ludii due to two key advantages: a large set of available games, and a graphical user interface available for running agents as well as human vs. AI games. The latter allows agents to print custom messages on the screen, alongside the game position, which is ideal for visualizing explanations.

The implemented MCTS variant can be configured by selecting which enhancements will be used for providing explanations.
Note that the difficulty of obtaining consistent human-friendly descriptions increases with the number of involved enhancements, as they often drastically change the algorithm's behavior, leading to more subcases.

Our current stage of implementation focuses on explaining the decision of an agent from the perspective of its current position and providing insights into the quality of the position itself, based on standard MCTS statistics, sibling comparison, and statistics of the aforementioned enhancements. 
The algorithm generates \emph{post-hoc}, \emph{model-agnostic} explanations containing a mix of \emph{Why?} and \emph{Why not?} arguments.
It relies on a rule-based system to select the most important and informative pieces of information, and a simple template-based system to provide final text in natural language.
Unlike most of the previous research, the algorithm and explanations are tailored to adversarial games.

Examples of interesting game situations accompanied by generated explanations have been shown in Table~\ref{tab:examples}. The examples try to cover particular scenarios where the enhancements provide utilizable additional knowledge, e.g., endgame with proven game result, GRAVE value overriding the standard move selection method. 
These examples show that adding this type of information may give more insight into why the search made a particular decision.

\section{Conclusion and Future Work}\label{sec:conclusion}

In this paper, we discuss a few particular aspects of Explainable Search, introduce and argue for the need for knowledge-agnostic explainability, and propose MCTS enhancements as a means to improve the quality of data gathered by MCTS.
Our work-in-progress solution has been developed as an agent within Ludii, featuring MCTS with customizable enhancements that display explanations in the \emph{Analysis} GUI tab during the game.
We have shown preliminary results in the form of explanations for specific game situations.
It is not the complete system, but shows the advantages provided by utilizing enhancements.

There are multiple future directions.
The first is to improve the quality of natural language by using Large Language Models as a post-processing tool. Although template-based explanations are simple to state, they sound very repetitive, too technical, and it is hard to properly manage their verbosity.
Our initial attempts in this regard, by including our generated explanations as part of the prompt, are promising but require further \emph{attention}.
The potential problem lies in incorporating calls to LLMs by agents running within the Ludii system.

The remaining steps are primarily focused on identifying additional methods for gathering game-agnostic knowledge and extracting more meaningful descriptions from this data.
Some ideas are categorizing MCTS tree shapes with respect to game tree shape and principal variation, including in explanations changes of in-game scores (if provided by the forward model), comparing position estimation to the result of null-move estimation (to detect zugzwang positions), and better formalization of traps and forced moves.

\bibliographystyle{IEEEtran} 
\bibliography{bibliography} 

\end{document}